\documentclass{article} 
\usepackage{latex_template,times}
\usepackage{hyperref}
\usepackage{url}
\usepackage{graphicx}
\usepackage{multirow}
\usepackage{adjustbox}
\usepackage{subcaption}
\usepackage{pgfplots}
\pgfplotsset{compat=1.14}
\usepackage{setspace}

\usepackage{authblk}
\author[1]{Pengwei Yang}
\author[1]{Chongyangzi Teng}
\author[1,2]{Jack George Mangos}
\affil[1]{Faculty of Computer Science, The University of Sydney, Sydney NSW 2000, Australia}
\affil[2]{Faculty of Medicine, University of New South Wales, Kensington NSW 2052, Australia}
\date{}                     
\title{Contaminated Images Recovery by Implementing Non-negative Matrix Factorisation}

\nipsfinalcopy 

\begin{document}


\maketitle

\begin{abstract}
Non-negative matrix factorisation (NMF) has been extensively applied to the problem of corrupted image data. Standard NMF approach minimises Euclidean distance between data matrix and factorised approximation. The traditional NMF technique is sensitive to outliers since it utilises the squared error of each data point, despite the fact that this method has proven effective. In this study, we theoretically examine the robustness of the traditional NMF, HCNMF, and $L 2,1$-NMF algorithms and execute sets of experiments to demonstrate the robustness on ORL and Extended YaleB datasets. Our research indicates that each algorithm requires a different number of iterations to converge. Due to the computational cost of these approaches, our final models, such as the HCNMF and $L 2,1$-NMF model, fail to converge within the iteration parameters of this work. Nonetheless, the experimental results illustrate, to some extent, the robustness of the aforementioned techniques. 
\end{abstract}

\section{Introduction}
\label{gen_inst}
The processing of, and automated learning from noisy image data is a pressing challenge in the field of machine learning. This is because there are an enormous number of sources of noisy image data which present bountiful opportunities for meaningful research. These include blurred hand-held photography for image reconstruction and denoising, or a number of medical image streams with intrinsic noise, such as speckle noise in ultrasonographic organ imaging, the high signal-to-noise ratio in the scattered vessel fragments against white and grey matter in cerebral CT, or time constraints (as are often faced in hospitals) limiting the quality of MRI data \cite{fan_image_1998} \cite{gupta_despeckling_2005}  \cite{wang_vc-net_2021} \cite{wu_learning-based_2019} \cite{haldar_regularised_2008}. As such, improving machine learning performance on noisy image datasets presents many opportunities for scientific and technological advancement.

Dimensionality reduction techniques can lead to “better predictions and more interpretable data” \cite{squires_non-negative_2019}. Non-negative matrix factorisation (NMF) is one group of methods used to address noisy image data. NMF methods involve the decomposition of multivariate data into smaller, or compressed, representative matrices \cite{lee_algorithms_2001}. In a data matrix consisting of multivariate n-dimensional data vectors, as below, the vectors are factorised into an n*r matrix W and an r*m matrix H, where r is typically smaller than both n and m. As such, each data vector v is approximated by a linear combination of the columns of W, weighted by the components of h:\,$	v \simeq Wh $.

As the number of basis vectors in W is generally substantially smaller than the number in the original data matrix, good approximation of the original data is achieved only if underlying latent structures are identifiable \cite{lee_algorithms_2001}. However, in cases where such underlying latent patterns are present, and where the non-negativity requirement is satisfied, NMF demonstrates good performance compared with other methods, such as PCA or vector quantization, particularly where PCA and NMF are used together vs. PCA alone \cite{lee_algorithms_2001} \cite{zhao_facial_2008}. 

This paper will test the robustness of the standard NMF, HCNMF and  $L_{2,1}$-NMF algorithms to an image clustering task on two related datasets. Both datasets are greyscale headshots – the aim being to cluster these into groupings based on the individual being photographed. The ORL “Dataset of Faces” dataset contains 400 images of 40 subjects, with variation in lighting, facial expression, and facial details (including glasses); the background, facial positioning, and image size is uniform \cite{att_laboratories_database_2001}. The Extended YaleB dataset is larger, containing 2414 images of 38 subjects; each subject is photographed in 9 different poses and under 64 illumination conditions. All images are manually aligned, and of uniform size \cite{noauthor_yale_nodate}. 

\section{Related Work}
\label{headings}
\subsection{Standard NMF}

NMF methods are used to capitalise on the non-negativity property of the datasets upon which they are used. As such, their use is limited to circumstances in which this property can be demonstrated, but they have previously been used in areas including text mining, hyper-spectral imaging, and genetic research \cite{guan_truncated_2019}. The standard algorithm is based on the decomposition of the original data matrix into two lower-dimensional non-negative matrix factors, where the Euclidean distance between the product and the original data matrix is used to optimise the solution\cite{guan_truncated_2019}. 

However, as the NMF algorithm only allows additive combinations, it is optimal when the dataset contains additive Gaussian noise, and therefore fails on highly corrupted or otherwise heterogeneously noisy datasets \cite{guan_truncated_2019}. This, importantly for our paper, includes datasets of faces where the faces are obscured by glasses or scarves. 

\subsection{HCNMF}

Many other variants of the NMF algorithm have been developed for different purposes. Hypersurface cost-based NMF, first proposed by Hamza and Brady, works by minimising the summation of hypersurface costs of errors. The benefit of the HCNMF method is that it produces an algorithm less sensitive to outliers than the $L2$-norm approach. 

The aforementioned  properties have meant the HCNMF function has been shown to outperform other methods (including NMF, PCA, cNMF, among others) in the factorisation of a spectral library \cite{hamza_reconstruction_2006}, and in selecting differentially expressed genes and tumour classification \cite{jiao_hyper-graph_2020}. However, one significant drawback to the HCNMF method is its computational demand. Because the algorithm utilises Armijo’s rule-based line search, optimisation is time-consuming \cite{guan_truncated_2019}.

\subsection{$L_{2,1}$-NMF}

The standard NMF algorithm employs a Frobenius norm-based loss function, which makes it sensitive to noise, as described above. In order to address this sensitivity, the $L_{2,1}$ method[14] employs a loss function whereby the $L_{2,1}$ norm of the error matrix is minimised; this reduces the influence of outliers/noisy data points by inhibiting their significance in learning the subspace \cite{guan_truncated_2019}. 

In reducing the sensitivity of the NMF algorithm to noisy data points, the $L_{2,1}$ method is more robust on heterogeneous datasets. As such, $L_{2,1}$-NMF can perform more efficiently than standard NMF when dealing with real-world data which contains noise. Furthermore, robust NMF such as $L_{2,1}$-NMF could be used for feature selection when applied to complex domains \cite{diaz2021analysis}.

\section{Methodology}
\label{others}

\subsection{Cost function and optimisation}
\subsubsection{Standard NMF}
The objective function of the standard NMF algorithm is defined as: 
 $$min_{W\geq0,H\geq0}||V-WH||_{F}^2 $$
where V is the set of input data vectors and W, H are the factor matrices. This objective function is typically solved using the multiplicative update rule (MUR) \cite{lee_algorithms_2001}.
The iterative updating algorithm of standard NMF is shown below:
$$F_jk\Leftarrow F_jk\frac{(XDG^T)_{jk}}{(FGDG^T)_{jk}}$$

$$G_ki\Leftarrow G_ki\frac{(F^TXD)_{ki}}{(F^TFGD)_{ki}}$$
\subsubsection{HCNMF}
The objective function of HCNMF is defined as:
$${{\Large \sum }_{ij}(\sqrt{1+(V-WH)^2_{ij}}-1)}$$
where the cost function is defined as:
$$\delta(x)=\sqrt{1+x^2}-1$$
The cost function is differentiable and bounded, and that the cost function is quadratic when the argument is small and linear when the argument is large (as shown in Fig.\ref{HCNMI_lossfunction}):

\begin{figure}[htb]
\begin{center}
\includegraphics[width=0.5\textwidth]{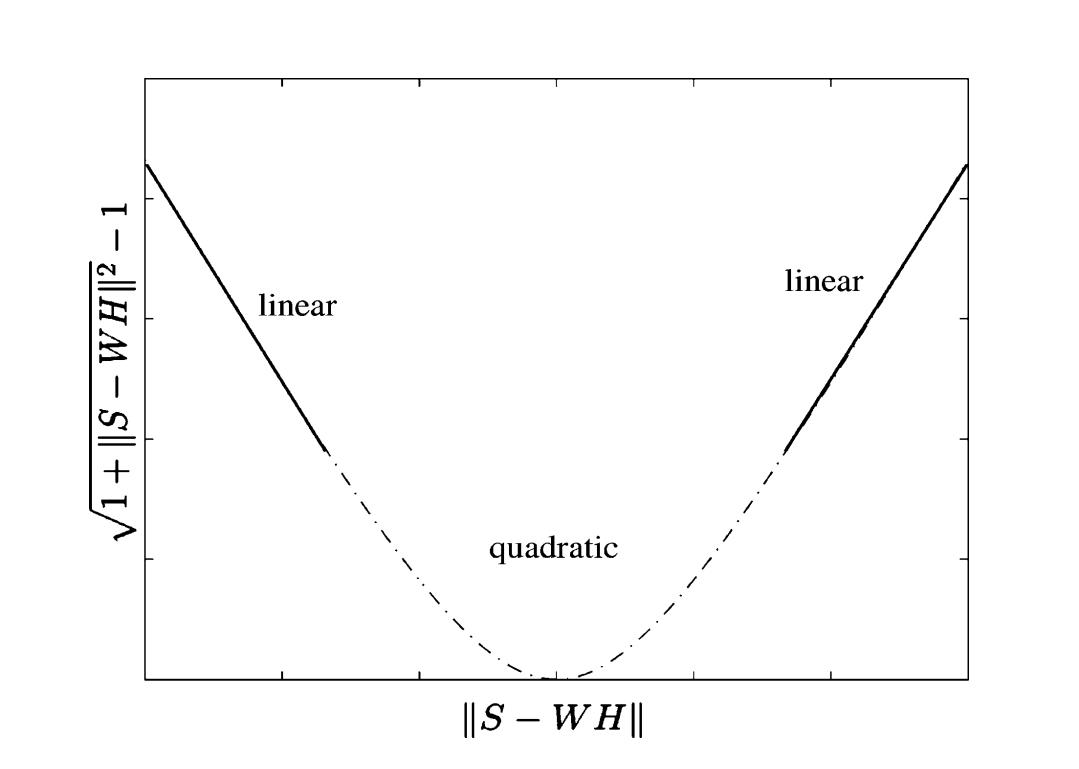}
\end{center}
\caption{\label{HCNMI_lossfunction}Loss Function of HCNMF \cite{hamza_reconstruction_2006}}
\end{figure}

The iterative updating algorithm of HCNMF is shown below:
$$W_{ik}^{(t+1)}=W_{ik}^{(t)}-\alpha_{ik}^{(t)}\frac{(WHH^T)_{ik}^{(t)}-(SH^T)_{ik}^{(t)}}{\sqrt{1+||S-WH||}}$$

$$H_{kj}^{(t+1)}=H_{kj}^{(t)}-\beta_{kj}^{(t)}\frac{(W^TWH)_{kj}^{(t)}-(W^TS)_{kj}^{(t)}}{\sqrt{1+||S-WH||}}$$
where alpha and beta are the step sizes opted at each iteration via Armijo rule for inexact line search \cite{hamza_reconstruction_2006}. Armijo rule is shown below \cite{1966Minimization}.
$$f(x_k-\alpha_{mk}\nabla f(x_k))-f(x_k)\leq-\frac{1}{2}\alpha_{mk}|\nabla f(x_k)|^2$$

\subsubsection{$L_{2,1}$-NMF}
The standard NMF utilises the Frobenius-norm-based loss function, which may mean that the the squared residual error might be large for any individual point. In order to handle that problem, $L_{2,1}$-NMF uses a robust formulation of the error function, which could be found as follows:
$$min_{W\geq0,H\geq0}||V-WH||_{2,1}$$
while the $L_{2,1}$-norm is defined as:
$$||E||_{2,1}={\Large \sum }_{j=1}^n||E_j||_2$$
This method no longer employs the squared error, and hence why the $L_{2,1}$-NMF has better robustness to noise than standard NMF. 

The iterative updating algorithm of standard $L_{2,1}$-NMF is shown below:
$$D_{ii}\Leftarrow\frac{1}{\sqrt{{\Large \sum }_{j=1}^p(X-FG)_{ji}^2}}=\frac{1}{||x_-Fg_i||}$$

$$F_{jk}\Leftarrow F_{jk}\frac{(XDG^T)_{jk}}{(FGDG^T)_{jk}}$$
$$G_{kj}\Leftarrow G_{kj}\frac{(F^TXD)_{ki}}{(F^TFGD)_{ki}}$$
Researchers add a weighted matrix regulariser to the $L_{2,1}$-NMF, which aims to incorporate these weights to suppress outliers \cite{kong_robust_2011}. D is calculated as a diagonal matrix.
\subsection{Noise}
\begin{figure}[htb]
\begin{center}
\includegraphics[width=1\textwidth]{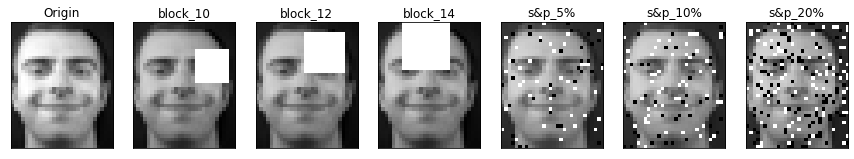}
\end{center}
\caption{\label{fig:Noise}Demonstration of both block-occlusion noise (images 2-4) and salt-and-pepper noise (images 5-7) on an example image from the ORL dataset}

\end{figure}

Real image data often contains noise. To compare how different NMF algorithms perform with different types of corruption, we simulated two typical types of noise: salt and pepper noise, and block-occlusion. We applied these types of noise to the two image datasets, ORL and Extended YaleB, to be used in our paper. Examples of these types of noise, including the varying degree to which they were applied, is shown in Fig.\ref{fig:Noise}. 

\subsubsection{Salt and Pepper (S\&P) Noise}
S\&P noise is also known as impulse noise, and is defined as a sparsely occurring white and black pixel distribution superimposed on images \cite{alqadi_salt_2018}. It can be caused by several reasons, including defective camera sensors, or software or hardware failure in image capturing, or due to transmission/conversion error \cite{alajlan_detail_2004}. This type of noise will contaminate the image by a factor between zero and the maximum pixel value,and will do so in a random, pixel-by-pixel distribution. Typically the intensity value for pepper noise is close to 0, and close to 255 for salt noise:
$$\eta(x,y)=\{^{0,\quad Pepper\,noise}_{255,\quad Salt\,noise}\}$$
Of note, S\&P noise can violate the Gaussian distribution noise assumption of standard NMF \cite{alqadi_salt_2018}. 

In this paper, S\&P noise is simulated by randomly substituting a proportion p of the total image pixel count for any one image with either grayscale value 0 (black) or 255 (white). We have varied the percentage of corrupted pixels (p) to be 5\%, 10\% and 20\% to demonstrate how the robustness of algorithms changes in response to higher amounts of noise. The proportion of salt to pepper noise is 0.5. 
\subsubsection{Block Occlusion}
Image occlusion is also a common form of noise that can depreciate the performance of image recognition. This is a challenging type of noise as, in contrast to the dispersed noise of S\&P, block occlusion obscures all information from one distinct region. This means that more traditional methods of overcoming S\&P noise, such as standard or adaptive median filtration, are ineffective \cite{sanchez_determining_2001}. This type of noise, for a similar reason, is also challenging for NMF methods, which need to lean more heavily on dictionary information to account for the large amount of contiguous missing information. 

In our paper, block occlusion is simulated by randomly placing a single b x b sized square block on each image. The block is given a uniform pixel greyscale value of 255. Three different block sizes were superimposed on distinct copies of the dataset, with b = 10, 12 and 14, to demonstrate how the robustness of tested algorithms changed with more information loss. The minimum block size and maximum block size imply 9\% and 18\% outlier for ORL images and 5\% and 10\% for Extended YaleB images.

\subsection{Evaluation Metrics}

We utilise three metrics for scoring algorithmic performance in each setting of noise.
\subsubsection{RRE}
The first is the relative reconstruction error (RRE). This metric measures the normalised distance between the clean dataset (Vhat) and the estimated dataset (WH) as an indication of the overall accuracy of the prediction, as below:
$$RRE=\frac{||\hat{V}-WH||_F}{||\hat{V}||_F}$$
\subsubsection{Accuracy}
The second is the average accuracy; this metric relates to the success of our algorithms in clustering the reconstructed images into their respective subject identities. This is calculated as the proportion of correctly clustered images compared to the overall sample size n, as below:
$$Acc(Y,Y_{pred})=\frac{1}{n}\,\sum_{i=1}^n1\{Y_{pred}(i)==Y(i)\}$$
\subsubsection{NMI}
Finally, normalised mutual information is used to evaluate clustering quality from the perspective of overall information gain, defined as the normalisation on the mutual information between cluster papers and the pre-existing input label. The normalisation used is the average of the entropy of the cluster paper and that of the input labelling\cite{kong_robust_2011}, as below:
$$NMI(Y,Y_{pred})=\frac{2*I(Y,Y_{pred})}{H(Y)+H(Y_{pred})}$$

\section{Experiment}
\subsection{Experimental Setup}

The experiments were performed on the ORL and Extended YaleB datasets. To reduce the computational complexity, we have resized all ORL images to 30x37 pixels and all ORL images to 42x48 pixels.

Each of the ORL and Extended YaleB datasets was exposed to the same set of noise - clean, block 10, 12, and 14, and S\&P 5, 10, and 20\%. Three algorithms were then assessed for their robustness - NMF, HCNMF, and the $L_{2,1}$-NMF. Each algorithm was tested on all proportions of noise for both datasets. 

For rigorous performance evaluation, we have repeated each experiment 5 times by randomly sampling 90\% of data from each whole dataset. The average and standard deviation of the metrics are reported. 
\subsection{Experimental Results}

\begin{figure}[htb]
\begin{center}
\includegraphics[width=1\textwidth]{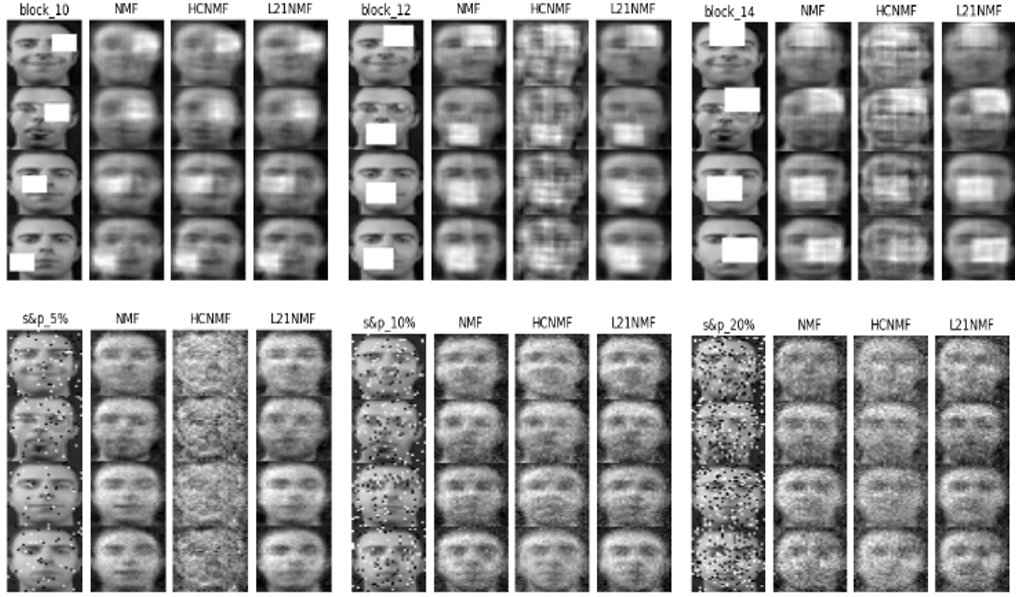}
\end{center}
\caption{\label{fig:Reconstructive image} Reconstructed image - From left to right: Polluted image; Results by NMF; Results by HCNMF; Results by $L_{2,1}$-NMF}
\end{figure}

Table~\ref{fig:Performance table} and Figure~\ref{fig:ORL Accuracy} to~\ref{fig:Yale NMI} shows the average performance score of standard NMF, HCNMF and L2,1 NMF for both the ORL dataset and the extended YaleB dataset. Figure\ref{fig:Reconstructive image} is a visual representation of the robustness of the models. The reconstructed images are produced by multiplication of matrices W and H that are learned based on the images with noise.

\begin{table}[htb]
\caption{\label{fig:Performance table}Relative reconstruction error\%, Accuracy\% and NMI\% with standard deviation}
\begin{subtable}{1\linewidth}\centering
\caption{ORL dataset}\label{tab:1a}
\begin{center}
\begin{adjustbox}{width=1\textwidth}
\begin{tabular}{|c||c|c|c||c|c|c||c|c|c|}
\hline\hline
\multirow{2}{*}{Noise Type} & \multicolumn{3}{c||}{RRE} & %
    \multicolumn{3}{c||}{ACC} & \multicolumn{3}{c|}{NMI}\\
\cline{2-10}
 & NMF & HCNMF & $L_{2,1}$ & NMF & HCNMF & $L_{2,1}$ & NMF & HCNMF & $L_{2,1}$ \\
\hline\hline
Clean & 12.59$\pm$0.001 & \textbf{12.38$\pm$0.018} & 90.97$\pm$0. & 71.56$\pm$0.026 & \textbf{76.94$\pm$0.026} & 73.89$\pm$0.018 & 84.22$\pm$0.012 & \textbf{87.36$\pm$0.013} & 85.18$\pm$0.007 \\
\hline
block 10 & \textbf{32.42$\pm$0.001} & 32.43$\pm$0.001 & 91.16$\pm$0. & 29.44$\pm$0.014 & 26.5$\pm$0.025 & \textbf{30.17$\pm$0.021} & \textbf{46.88$\pm$0.008} & 43.42$\pm$0.023 & 46.79$\pm$0.024 \\
\hline
block 12 & \textbf{38.49$\pm$0.001} & 38.6$\pm$0.002 & 91.23$\pm$0. & 22.72$\pm$0.008 & \textbf{23.83$\pm$0.013} & 22.94$\pm$0.004 & 41.45$\pm$0.01 & \textbf{41.59$\pm$0.004} & 40.32$\pm$0.007 \\
\hline
block 14 & \textbf{43.55$\pm$0.002} & 43.59$\pm$0.002 & 91.24$\pm$0. & 22.67$\pm$0.015 & \textbf{24.06$\pm$0.016} & 21.67$\pm$0.006 & 40.58$\pm$0.015 & \textbf{41.23$\pm$0.015} & 40.83$\pm$0.007 \\
\hline
S\&P 5\% & \textbf{15.63$\pm$0.} & 22.36$\pm$0.206 & 91.07$\pm$0. & 69.83$\pm$0.023 & 54.78$\pm$0.073 & \textbf{70.89$\pm$0.016} & 82.26$\pm$0.014 & 65.35$\pm$0.045 & \textbf{82.35$\pm$0.011} \\
\hline
S\&P 10\% & \textbf{18.8$\pm$0.001} & 19.92$\pm$0.001 & 91.15$\pm$0. & 65.17$\pm$0.022 & \textbf{69.83$\pm$0.015} & 66.28$\pm$0.023 & 78.56$\pm$0.007 & \textbf{81.83$\pm$0.013} & 78.83$\pm$0.015 \\
\hline
S\&P 20\% & \textbf{24.22$\pm$0.001} & 26.31$\pm$0.001 & 91.32$\pm$0. & 52.44$\pm$0.015 & \textbf{54.06$\pm$0.02} & 52.28$\pm$0.019 & 68.17$\pm$0.006 & \textbf{71.27$\pm$0.015} & 68.36$\pm$0.018 \\
\hline

\end{tabular}
\end{adjustbox}
\end{center}
\caption{Extended YaleB dataset}\label{tab:2}
\begin{center}
\begin{adjustbox}{width=1\textwidth}
\begin{tabular}{|c||c|c|c||c|c|c||c|c|c|}
\hline\hline
\multirow{2}{*}{Noise Type} & \multicolumn{3}{c||}{RRE} & %
    \multicolumn{3}{c||}{ACC} & \multicolumn{3}{c|}{NMI}\\
\cline{2-10}
 & NMF & HCNMF & $L_{2,1}$ & NMF & HCNMF & $L_{2,1}$ & NMF & HCNMF & $L_{2,1}$ \\
\hline\hline
Clean & 18.55$\pm$0.001) & 78.37$\pm$0. & 81.59$\pm$0.002 & 24.29$\pm$0.011 & 9.24$\pm$0.003 & 18.56$\pm$0.006 & 32.58$\pm$0.016 & 9.6$\pm$0.006 & 25.59$\pm$0.009 \\
\hline
block10 & 45.14$\pm$0.001 & 79.25$\pm$0. & 86.59$\pm$0. & 11.42$\pm$0.008 & 9.17$\pm$0.002 & 9.61$\pm$0.002 & 13.43$\pm$0.017 & 8.83$\pm$0.008 & 10.66$\pm$0.008 \\
\hline
block12 & 53.72$\pm$0.001 & 79.53$\pm$0. & 87.26$\pm$0. & 10.43$\pm$0.009 & 9.$\pm$0.002 & 9.48$\pm$0.003 & 11.88$\pm$0.012 & 9.18$\pm$0.002 & 9.53$\pm$0.005 \\
\hline
block14 & 62.59$\pm$0.002 & 79.8$\pm$0. & 87.9$\pm$0. & 9.81$\pm$0.003 & 9.16$\pm$0.003 & 9.11$\pm$0.002 & 11.08$\pm$0.007 & 8.74$\pm$0.004 & 9.17$\pm$0.004 \\
\hline
S\&P5\% & 19.86$\pm$0.001 & 78.88$\pm$0.001 & 85.14$\pm$0. & 23.51$\pm$0.01 & 9.36$\pm$0.004 & 17.88$\pm$0.006 & 31.55$\pm$0.008 & 9.6$\pm$0.003 & 24.21$\pm$0.016 \\
\hline
S\&P10\% & 21.86$\pm$0. & 79.21$\pm$0.001 & 86.29$\pm$0.001 & 23.24$\pm$0.008 & 9.27$\pm$0.001 & 16.94$\pm$0.009 & 31.39$\pm$0.013 & 9.47$\pm$0.006 & 22.93$\pm$0.014 \\
\hline
S\&P20\% & 26.99$\pm$0.001 & 79.74$\pm$0. & 87.57$\pm$0. & 21.61$\pm$0.012 & 9.01$\pm$0.001 & 14.$\pm$0.005 & 29.5$\pm$0.012 & 9.28$\pm$0.004 & 18.22$\pm$0.007 \\
\hline

\end{tabular}
\end{adjustbox}
\end{center}
\end{subtable}
\end{table}

\begin{figure}[htb]
\begin{minipage}[b]{0.4\textwidth}
\begin{tikzpicture}
    \begin{axis}
    [
   width=0.8\textwidth,
   scale only axis,
   xticklabel style = {rotate=30,anchor=east},
   xticklabels from table={ORL_Acc.dat}{Noise},xtick=data,
   ymin=0, ymax=100,
   ylabel={Average Accuracy (\%)},
   axis lines*=left,
   legend style ={ at={(0.8,0.3)}, 
        anchor=north west, draw=black, 
        fill=white,align=left,
        nodes={scale=0.5, transform shape}},
    cycle list name=black white,
]
\addplot[black] table [y= NMF,x=X]{ORL_Acc.dat};
\addlegendentry{NMF}
\addplot[orange,thick,mark=square*] table [y= HCNMF,x=X]{ORL_Acc.dat};
\addlegendentry{HCNMF}
\addplot[blue,densely dotted,mark=triangle*] table [y= L21NMF,x=X]{ORL_Acc.dat};
\addlegendentry{L21NMF}
\end{axis}
\end{tikzpicture}
\caption{\label{fig:ORL Accuracy}ORL - Accuracy}
\end{minipage}
\hfill
\begin{minipage}[b]{0.4\textwidth}
\begin{tikzpicture}
    \begin{axis}
    [
   width=0.8\textwidth,
   scale only axis,
   xticklabel style = {rotate=30,anchor=east},
   xticklabels from table={ORL_NMI.dat}{Noise},xtick=data,
   ymin=0, ymax=100,
   ylabel={Normalised Mutual Information (\%)},
   axis lines*=left,
   legend style ={ at={(0.8,0.3)}, 
        anchor=north west, draw=black, 
        fill=white,align=left,
        nodes={scale=0.5, transform shape}},
    cycle list name=black white,
]
\addplot[black] table [y= NMF,x=X]{ORL_NMI.dat};
\addlegendentry{NMF}
\addplot[orange,thick,mark=square*] table [y= HCNMF,x=X]{ORL_NMI.dat};
\addlegendentry{HCNMF}
\addplot[blue,densely dotted,mark=triangle*] table [y= L21NMF,x=X]{ORL_NMI.dat};
\addlegendentry{L21NMF}
\end{axis}
\end{tikzpicture}
\caption{\label{fig:ORL NMI}ORL - NMI}
\end{minipage}
\end{figure}

\begin{figure}[htb]
\begin{minipage}[b]{0.4\textwidth}
\begin{tikzpicture}
    \begin{axis}
    [
   width=0.8\textwidth,
   scale only axis,
   xticklabel style = {rotate=30,anchor=east},
   xticklabels from table={Yale_Acc.dat}{Noise},xtick=data,
   ymin=0, ymax=60,
   ylabel={Average Accuracy (\%)},
   axis lines*=left,
   legend style ={ at={(0.7,1)}, 
        anchor=north west, draw=black, 
        fill=white,align=left,
        nodes={scale=0.5, transform shape}},
    cycle list name=black white,
]
\addplot[black] table [y= NMF,x=X]{Yale_Acc.dat};
\addlegendentry{NMF}
\addplot[orange,thick,mark=square*] table [y= HCNMF,x=X]{Yale_Acc.dat};
\addlegendentry{HCNMF}
\addplot[blue,densely dotted,mark=triangle*] table [y= L21NMF,x=X]{Yale_Acc.dat};
\addlegendentry{L21NMF}
\end{axis}
\end{tikzpicture}
\caption{\label{fig:Yale Accuracy}Extended YaleB - Accuracy}
\end{minipage}
\hfill
\begin{minipage}[b]{0.4\textwidth}
\begin{tikzpicture}
    \begin{axis}
    [
   width=0.8\textwidth,
   scale only axis,
   xticklabel style = {rotate=30,anchor=east},
   xticklabels from table={Yale_NMI.dat}{Noise},xtick=data,
   ymin=0, ymax=60,
   ylabel={Normalised Mutual Information (\%)},
   axis lines*=left,
   legend style ={ at={(0.7,1)}, 
        anchor=north west, draw=black, 
        fill=white,align=left,
        nodes={scale=0.5, transform shape}},
    cycle list name=black white,
]
\addplot[black] table [y= NMF,x=X]{Yale_NMI.dat};
\addlegendentry{NMF}
\addplot[orange,thick,mark=square*] table [y= HCNMF,x=X]{Yale_NMI.dat};
\addlegendentry{HCNMF}
\addplot[blue,densely dotted,mark=triangle*] table [y= L21NMF,x=X]{Yale_NMI.dat};
\addlegendentry{L21NMF}
\end{axis}
\end{tikzpicture}
\caption{\label{fig:Yale NMI} Extended YaleB - NMI}
\end{minipage}
\end{figure}

\subsubsection{ORL dataset}

All three models have similar performance. The accuracy and NMI on clean images are around 74\% and 86\% respectively. When we added salt\&pepper noise, all three models still have satisfactory performance at 5\% S\&P corruption. As the corruption percentage increases, a decrease in the performance for all three models can be observed. For S\&P 10\% and 20\%, HCNMF is slightly more robust. HCNMF has the worst performance at 5\% corruption; this may be due to volatility, as the experiment is only repeated 5 times and outlier results could significantly skew the average score. This is evidenced by the higher standard deviation observed in table~\ref{fig:Performance table}.

All three models have, unsurprisingly (given the challenging nature of the noise), a significant deterioration in performance with block occlusion. For a 10*10 block, which implies a 9\% proportion of outliers for ORL images, the performance is significantly worse compared with a roughly equivalent salt and pepper noise of 10\%.

The standard NMF converged before reaching the maximum iteration of 5000, however both HCNMF and $L_{2,1}$-NMF fail to converge. The results reported for HCNMF and $L_{2,1}$-NMF are based on their results at the max iteration steps. 

 \subsubsection{Extended YaleB dataset}

This dataset contains 2414 images and requires a significantly longer time to run an experiment. The optimisation of HCNMF and $L_{2,1}$-NMF is more difficult and time-consuming compared to standard NMF. Due to the limited computational resources available, our team chose to cap the iteration steps at 3000 for HCNMF and 100 for $L_{2,1}$-NMF. At the max iteration both models have not converged to error bound, which has significantly impacted each respective model's performance.

When the iteration steps/running time is restricted, from the results table and chart, we can see the standard NMF outperformed both HCNMF and $L_{2,1}$-NMF.

Similar to what we observed from ORL data results, as image corruption increases the performance of the three NMF methods decreases. All three models perform better with respect to salt and pepper noise compared to block occlusion.

\subsection{Reflection for future improvement}
For future work, we would like to make a few improvements in the experiment design.

Firstly, increasing the max iteration cap to allow HCNMF and $L_{2,1}$ method to converge would potentially demonstrate their better performance on the two datasets. However, we were unable to facilitate this amount of testing due to limitations in time available and computational resources. As such, the reported results for these two methods are unconverged results. This has significantly impacted their performance and we are unable to fairly compare the robustness of the models relatively to each other.

To demonstrate that increasing the number of iterations would improve the performance when it gets closer to convergence point (global minimum point of the cost function), our team tested only the L2,1-NMF model using only 20\% Salt and Pepper noise and a higher iteration step threshold. At the baseline number of iterations, the model does not converge with H and W updates differences above the threshold of 1e5 and continuing to decrease, however, as can be seen in Table 2, if the number of iteration steps is increased, the model performance demonstrates an improvement accordingly. Further experiments would require significantly longer run time due to the 5-fold cross-validation we’ve implemented, to trade off performance and run-time, we have set a lower threshold.

We would also like to test on a wider range of noise corruption. In this experiment, our team only tested three values for each type of noise. We have not compared the performance on extreme outliers, for example where more than 50\% of the pixels are corrupted. In addition, due to limited points tested, we are not able to observe the rate of the performance deterioration and at which points the method starts to deteriorate significantly.

Lastly, to further improve the rigorousness of the performance evaluation, we would ideally repeat each experiment more than five times. However, increasing the number of repeate experiments will further increase the computational time.

\begin{table}[t]
\caption{Iteration impact demonstration}
\label{iteration demonstration}
\begin{center}
\begin{adjustbox}{width=1\textwidth}
\begin{tabular}{|c||c|c|c|c|c|c|}
\hline
 Step & Update diff of W & Update diff of H & Training time (mins) & RRE \% & Accuracy
\% & NMI \% \\
\hline\hline
1000& 0.0366	&0.0705	&0.87	&91.3	&53.0	&68.4\\
5000& 0.0362	&0.0736	&4.16	&91.3	&48.5	&65.8\\
10000& 0.0359	&0.0748	&8.17	&91.3	&49.0	&66.1\\
20000& 0.0358	&0.0752	&16.35	&91.3	&56.5	&71.0\\
\hline
\end{tabular}
\end{adjustbox}
\end{center}
\end{table}

\section{Conclusion and Future work}
\label{others}
In this paper, we implemented three types of NMF algorithms, namely standard NMF, HCNMF and $L_{2,1}$-NMF. We theoretically analysed the robustness of HCNMF and $L_{2,1}$-NMF. Considering the computational complexity of HCNMF and $L_{2,1}$-NMF methods, we chose to limit the number of update iterations, which meant that the models did not converge to the error bound at the cut-off point. As a result, our experiments have limited ability to demonstrate the robustness of both HCNMF and $L_{2,1}$-NMF. In the future, we could do further experiments to discover the range of convergence when establishing various models with ORL and YaleB dataset. 

\cite{10.5555/3454287.3454901}

\bibliographystyle{splncs04}
\bibliography{main_content.bib}

\end{document}